%% file: main.tex
\pdfoutput=1
\documentclass[11pt]{article}

\usepackage[preprint]{acl}

\usepackage{times}
\usepackage{latexsym}
\usepackage{amsmath}
\usepackage{float}
\usepackage{array}
\usepackage{textgreek}
\usepackage[T1]{fontenc}
\usepackage[utf8]{inputenc}

\usepackage{microtype}

\usepackage{inconsolata}

\usepackage{graphicx}

\newif\ifcomment
\commenttrue
\commentfalse
\ifcomment
\newcommand{\shirin}[1]{{\bf \textcolor{purple}{Shirin: #1}}}
\newcommand{\poojitha}[1]{{\bf \textcolor{blue}{Poojitha: #1}}}
\else
\newcommand{\shirin}[1]{}
\newcommand{\poojitha}[1]{}
\fi

\title{Attacks against Abstractive Text Summarization Models through \\ Lead Bias and Influence Functions}

\author{
 \textbf{Poojitha Thota}, \textbf{Shirin Nilizadeh} \\
 Department of Computer Science and Engineering \\
 The University of Texas at Arlington \\
 \small{
   \href{mailto:poojitha.thota@mavs.uta.edu}{poojitha.thota@mavs.uta.edu}, 
   \href{mailto:shirin.nilizadeh@uta.edu}{shirin.nilizadeh@uta.edu}
 }
}

\begin{document}
\maketitle
\begin{abstract}
Large Language Models (LLMs) have introduced novel opportunities for text comprehension and generation. Yet, they are vulnerable to adversarial perturbations and data poisoning attacks, particularly in tasks like text classification and translation. However, the adversarial robustness of \emph{abstractive text summarization} models remains less explored. In this work, we unveil a novel approach by exploiting the inherent lead bias in summarization models, to perform adversarial perturbations. Furthermore, we introduce an innovative application of influence functions, to execute data poisoning, which compromises the model's integrity. This approach not only shows a skew in the models' behavior to produce desired outcomes but also shows a new behavioral change, where models under attack tend to generate \emph{extractive} summaries rather than \emph{abstractive} summaries. 
\end{abstract}

\input{introduction}

\input{related-work}
\input{threat_model}

\input{methodology_advperturbations}

\input{methodology_poisoning}

\input{experimental-setup}

\input{Evaluation}

% \newpage 
\input{conclusion}
\input{discussion}
\input{ethics}

\bibliography{main}
\input{Appendix}

\end{document}

%% file: introduction.tex
\section{Introduction}
%-------------------------------------------------------------------------------
In recent years, with the advent of Large Language Models (LLMs), such as BERT~\cite{devlin2018bert}, BART~\cite{lewis2019bart}, T5~\cite{raffel2020exploring}, and GPT~\cite{radford2018improving,radford2019language}, the field of Natural Language Processing (NLP) has witnessed a monumental transformation. 
These models have revolutionized the way how machines understand and generate human language, offering capabilities in a wide range of applications from text classification, machine translation, and question-answering to text summarization. 
In particular, text summarization benefits from LLMs to consume vast amounts of information and provide concise and coherent summaries.

However, LLM's susceptibility towards adversarial tactics and poisoning attacks presents a critical vulnerability. 
Attacks mainly involve making subtle modifications to the model's input to produce incorrect or misleading outputs~\cite{ebrahimi2017hotflip}.
To date, studies have shed light on how adversarial inputs can impact models performing the task of text classification and translation~\cite{garg2020bae}. 
Recent works have started to study the impact of adversarial perturbations on text summarization. For instance, they have shown that minor adversarial perturbations like synonym substitution~\cite{chen2023improving} or utilizing homoglyphs~\cite{boucher2023boosting} can lower the quality of generated summaries. 

However, existing attack strategies developed for classification or translation tasks are not directly applicable to summarization due to differences in their goals and evaluation metrics. For example, while a change in predicted labels might measure success in classification, summarization requires consideration of more nuanced aspects, such as the quality, coherence, and context of the generated summaries. 
In addition, to the best of our knowledge, no work has systematically explored adversarial vulnerabilities specific to summarization tasks, especially in leveraging the LLMs and algorithmic properties. 
We employ a systematic, large-scale, layered approach that spans different levels ranging from characters, words, sentences, and documents. This comprehensive strategy allows us to explore a wider range of vulnerabilities specific to summarization models. 

Building on these methodological differences, we investigate exploiting lead bias~\cite{nallapati2017summarunner,grenander2019countering} within LLMs used for Text Summarization, which is the tendency of models to overly rely on the initial sentences of a document while generating summaries. 
We demonstrate how this bias poses a critical vulnerability in how text summarization models process and prioritize content. 
By embedding various types of adversarial perturbations to these leading sentences, we uncover a significant discrepancy in the model’s ability to present essential information accurately. 

Furthermore, poisoning attacks, where the training data is manipulated to degrade the model’s performance, have been explored for the tasks of text classification and translation~\cite{xu2021targeted,cui2022unified}. 
However, they are unexplored in the case of text summarization. 
Our work parallels dirty label attacks, a subset of poisoning attacks in which labels are intentionally altered to deceive models. 
We apply similar principles and implement new types of attacks specific to text summarization, where summaries change to contrastive or include toxic content without changing the training document's actual context or keywords. 

Central to our methodology is the innovative application of influence functions to strategically introduce poisoned data into the training dataset. Traditionally, these influence functions are used to assess the impact of a single data point on the overall model’s predictions~\cite{han2020explaining}. 
Leveraging these functions, we identify influential data points in the training dataset whose alteration can result in a modification in the behavior of these models. 
Moreover, we unveil a novel observation: The poisoned models tend to generate extractive summaries instead of abstractive summaries. 
This behavioral shift signifies not just a vulnerability to data poisoning attacks but also a fundamental alteration in how models process and summarize textual information under adversarial influence. 

This study examines Multi-Document Text Summarization (MDTS), which better simulates the information-gathering process in GenAI systems. These systems typically summarize information from multiple sources to answer user queries on specific topics. It also provides a more practical threat model, where the adversary modifies a few documents from various sources, potentially affecting the summarization outcome.
By systematically exposing these vulnerabilities in MDTS models, our work aims to motivate and inform future research into developing more secure and robust text summarization models that can maintain their integrity and performance in the face of potential adversarial manipulation.

The primary contributions of the work are as follows: 
\textbf{Comprehensive Evaluation of Adversarial Perturbations:} We analyze the response of text summarization models like BART, T5, and Pegasus, and the latest Chatbots, ChatGPT-3.5, Claude-Sonet, and Gemini to adversarial perturbations, ranging from character-level changes to broader manipulations at the word, sentence, and document level. 
\textbf{Lead Bias Exploitation Analysis:} We present the first study to exploit the lead bias in text summarization models for adversarial purposes, demonstrating a key vulnerability in model integrity. 
\textbf{Poisoning Attack Strategies during Model Fine-Tuning:} 
Using influence functions, we identify influential data points to poison training datasets, revealing a skew in the model's behavior and a shift in the model's tendency to generate extractive summaries instead of abstractive summaries when poisoned.
Our codes are available here: https://github.com/Rog11/summary-attack.

%% file: related-work.tex
\section{Related Work} 
\textbf{Multidocument Text Summarization.} 
Multi document text summarization involves synthesizing information from multiple text documents into a coherent and concise summary~\cite{mani2018multi}. 
Algorithms like TextRank~\cite{mihalcea-tarau-2004-textrank} and LexRank~\cite{Erkan_2004}, are some of the \emph{extractive} algorithms.
With the evolution of deep learning, more sophisticated \emph{abstractive methods} emerged, particularly those based on the transformer architecture, such as BART~\cite{lewis2019bart}, T5~\cite{raffel2020exploring}, PEGASUS~\cite{zhang2020pegasus}, etc.
These models utilize attention mechanisms and contextual embeddings to generate new text that replicate human-like narrative structures~\cite{zheng2020topic}

\textbf{Attacks in NLP.}  
Several works have studied the robustness of text classification tasks against adversarial inputs. 
The \emph{word-level techniques}, including HotFlip~\cite{ebrahimi2017hotflip}, TextFooler~\cite{jin2020bert}, and SemAttack~\cite{wang2022semattack} all produce subtle changes to the input text that lead the model to label the documents incorrectly. 
Many attacks are \emph{character-based}~\cite{madry2017towards,kurakin2018adversarial}. 
The well-known Fast Gradient Sign Method (FGSM)~\cite{goodfellow2014explaining} computes the gradient of the loss function with respect to the input. 
\emph{Sentence-based attacks} like sentence creation using gradient-based perturbation~\cite{hsieh2019natural} and Seq2seq Stacked Auto-Encoder~\cite{li2023efficiently} 
also produce adversarial inputs for text classification, aiming to preserve the general meaning of sentences.   
 
\textbf{Data Poisoning Attacks.} 
\label{subsec:data_poisoning}
Data Poisoning attacks are aimed at integrity of ML models, where attacker intentionally adds examples to training set to manipulate the behavior of the model at test time~\cite{shafahi2018poison}. 
These attacks in literature mainly include label-flipping or dirty label attacks~\cite{xiao2012adversarial}, where adversaries can manipulate the labels of training data points, to degrade the model's performance. 
Other types of these attacks include backdoor attacks~\cite{chen2017targeted}, which causes models to deviate from expected behavior when a trigger is encountered.  

%% file: threat_model.tex
\section{Threat Model}
\textbf{Adversarial Perturbations:} 
Adversaries can be motivated to perturb text summarization inputs during inference time so that they generate biased or misleading summaries. In this work, we assume the attacker's goal is to successfully implement \emph{sentence exclusion attack} to fool the summarization model not to use a specific sentence, here the lead sentence.
As a consequence of this attack, the model's output may suffer from \emph{degradation in quality}, i.e., generating incomplete, incoherent, or misleading summaries.  
For example, recently, summarization has been increasingly proposed to improve fact-checking processes~\cite{kazemi2021extractive,bhatnagar2022harnessing,yang2021scalable, hanikova2024towards}. Moreover, some practical implementations of fact-checkers by~\cite{Reuters} and~\cite{Google} utilize summarization to efficiently process and present fact-checked information to the public.
Our threat model considers an adversary who strategically implants fabricated news across multiple foreign outlets using an adversarial perturbation attack, ensuring they do not surface in the fact-checker platform's summaries. Consequently, misinformation evades debunking and persists in its spread.
We also assume a black box setting in which attackers do not have access to model parameters or training data.

\textbf{Data Poisoning Attacks:} We assume adversaries try to manipulate training data or release poisoned datasets into the public domain to poison the models that are later trained on this data, aiming to spread malicious behavior across a wide range of downstream applications. 
Adversaries can curate a dataset that appears legitimate but contains poisoned samples designed to gradually shift the behavior of the model toward the attacker's desired outcomes, including: 
(1) \emph{Sentiment inversion} to fool the summarization algorithm to flip the sentiment of a specific sentence in the output summary. 
(2) \emph{Toxic content inclusion} where the summarization algorithm or model is manipulated to incorporate toxic content into their generated summaries.
(3) \emph{Model behavioral change,} where the poisoned summarization model does not act as an abstractive algorithm, and instead of generating the summary, it extracts the exact sentences from the inputs. 
These are white-box attacks and the attacker requires a few high-performance GPUs in order to fine-tune the models and understand the influential data points, responsible for learning.

%% file: methodology_advperturbations.tex
\section{Adversarial Perturbations}
With their success on text classification, we examine the robustness of summarization models against  adversarial perturbations, which can be in different levels – character, word, sentence, and document. 
The space of possible modifications at every level is huge~\cite{ebrahimi2017hotflip}. We show how an attacker, leveraging the biases in summarization models, can implement \emph{sentence exclusion attack}, which can also result in \emph{quality degradation.}  

\begin{figure*}[ht!]
\centering
  \includegraphics[width=0.9\textwidth]{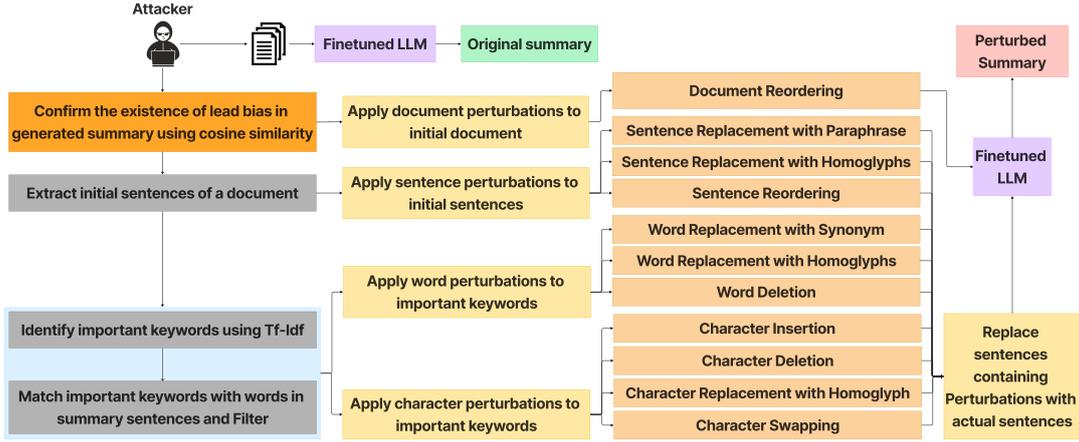}
\caption{Framework showing implementation of adversarial perturbations}
  \label{fig:adversarial_perturbations_framework}
\end{figure*}

In MDTS, models exhibit a phenomenon known as \emph{lead bias}, where they disproportionately focus on the initial sentences of a document~\cite{nenkova-etal-2011-automatic}. 
This bias arises due to training patterns where crucial information is typically located at the beginning of multiple documents. 
Additionally, \emph{document ordering bias} can play a role where models giving more weight to the content of documents presented earlier in the sequence~\cite{ravaut2023context}. 
We hypothesize that these biases make text summarization models vulnerable to adversarial perturbations. 
As shown in Figure~\ref{fig:adversarial_perturbations_framework}, we implemented eleven attacks, including four attacks using \emph{character-level} perturbations, three attacks using \emph{word-level} and \emph{sentence-level} perturbations, and one attack at the \emph{document level}. 

\textbf{Model fine-tuning and bias confirmation:} 
We verify the existence of lead bias in LLM-based text summarization models using publicly available pre-trained models and multi-document datasets. 
The models' susceptibility to lead and document ordering biases gives attackers a cue on where to modify the input documents to manipulate the summary. 
This can reduce the search space and efficiently influence the overall summary. Next, we formalize the adversarial perturbations and describe the process of identifying influential tokens. 

\textbf{Adversarial Perturbations Formalization:} 
For a set of documents \{$D_1, D_2, ..., D_k$\}, where each $D_{i}$ consists of sentences \{$s_{i1}, s_{i2}, ..., s_{in}$\}, we specifically target the lead sentences of the first document, $D_{lead}=\{s_{11}, s_{12}, ..., s_{1m}\}$, with $m$ being a small number, such as 2 or 3.
This targeted approach stems from the hypothesis that alterations in the lead sentences of the first document can disproportionately influence the overall summary. 

\textbf{Identification of important tokens:} 
In \emph{character} and \emph{word} level, we employ TF-IDF to determine the important words within $D_{lead}$. 
Instead of applying adversarial perturbations to all the important words in the set, we match the words present in sentences of summary and filter them to apply perturbations. 
This set of selected words is denoted as $W_{imp}$. 
Our adversarial strategy involves applying a perturbation function \textit{p} to $W_{imp}$. 
This function \textit{p(w)} is designed to apply perturbations across characters and words in the set of $W_{imp}$, encompassing insertions, deletions, or homoglyph, synonym replacements while adhering to the constraint of minimal perturbation. 
At the \emph{sentence level}, \textit{p(w)} is designed to apply perturbations across $D_{lead}$, encompassing replacement with paraphrases and homoglyphs and re-ordering.  
At the \emph{document level}, \textit{p(w)} is designed to apply perturbations across $D_1$ by changing the document's location from top to bottom.  
The application of \textit{p(w)} to $D_{lead}$ results in a perturbed version, $D_{lead}^{'}$. 
Table~\ref{tab:perturbations_examples_changes} in the Appendix shows examples, where the original sentence is \textit{``Anissa Weier is brought into court for a hearing last month}.''

\textbf{Character Swapping, Deletion and Insertion:} 
These perturbations can simulate common typo errors and input noise that can occur in real-world scenarios. 
We assess models' ability to correct or accommodate such variations in summarization. 

\textbf{Replacement with Homoglyphs:} 
Homoglyphs are visually similar characters/ words that are less noticeable to human readers and can be used for deceptive purposes. 
We assess models' adversarial robustness when one character or word at a time is replaced with its homoglyph counterpart. 

\textbf{Word Deletion:}
Important words or entities may be missing due to user input errors, censorship, or data corruption. 
We evaluate the models' ability to handle such missing information. 

\textbf{Word Replacement with Synonyms:}
Words can be expressed in multiple ways using synonyms. Motivated by the success of synonym replacement in attacking \emph{text classification} tasks, we test the models' ability to understand contextually equivalent expressions during summarization when one word at a time is replaced with its synonym.

\textbf{Sentence/Document Reordering:}
The order of sentences and paragraphs helps understand their context. We evaluate the models robustness against such changes in structure by moving 
one of the sentences in a document from the top to the bottom and placing the top document at the bottom. 

\textbf{Sentence Paraphrasing:} 
Models should be able to handle paraphrased expressions while capturing the core meaning. 
We test the models' ability to summarize effectively while replacing the original sentence with its paraphrased version.

%% file: methodology_poisoning.tex
\section{Influence Functions for Data Poisoning}
\label{sec:poisoning_methodology} 
The methodology we implemented for data poisoning is similar to dirty label attacks, which have proved to be successful in the case of text classification~\cite{xiao2012adversarial,shafahi2018poison}. However, 
these approaches are not directly applicable to text summarization. Specifically, text classification tasks involve labels that can be manipulated for a dirty label attack, where incorrect labels are intentionally introduced to degrade model performance. In contrast, text summarization does not rely on such labels, and it involves generating coherent summaries, where a different approach is required for data poisoning.
We propose a novel attack strategy tailored to Text Summarization models, where attackers can employ influence functions to systematically target and modify training data. Influence functions allow us to quantify the impact of a single data point on the model's predictions~\cite{cook1980characterizations}. By leveraging this information, attackers can identify the most influential training samples and strategically perturb them to manipulate the model's behavior. Our proposed approach differs from dirty label attacks in two key aspects. Firstly, instead of modifying labels, we focus on perturbing the content of summaries in training instances to either a contrastive or a toxic version.
Second, we utilize the influence functions to guide the selection of instances to be modified, making sure that the perturbations have a significant impact on the model's predictions. 

The framework to execute this attack is outlined in Figure~\ref{fig:poisoning_framework}, with the following components:  
\begin{figure}[t]
\centering
  \includegraphics[width=0.9\columnwidth]{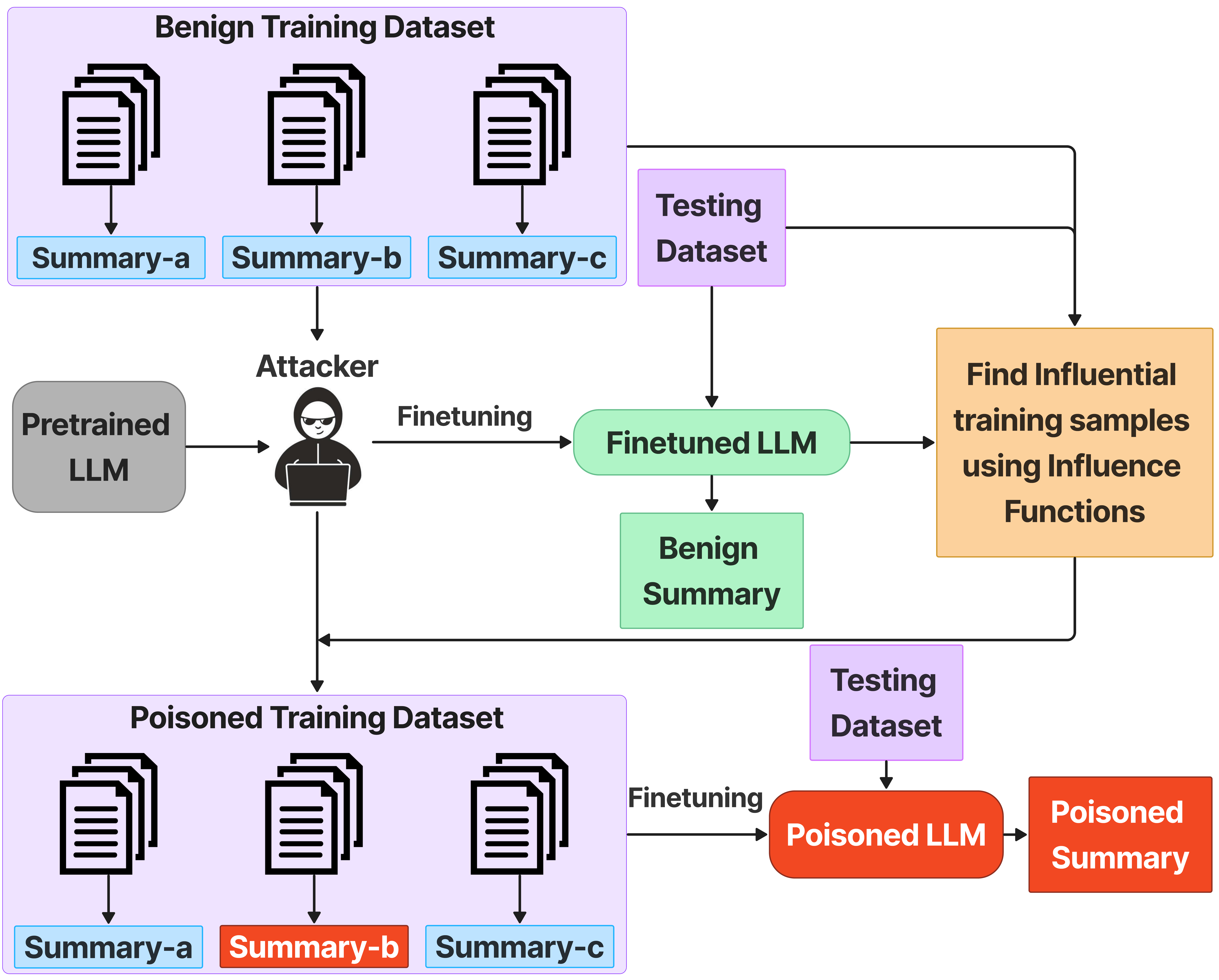}
\caption{Poisoning attack using Influence Functions}
  \label{fig:poisoning_framework}
\end{figure} 
\textbf{(1)~Initial setup:}
Initially, an attacker has access to a benign training dataset, a testing dataset, and a publicly available pre-trained LLM. 
The pre-trained LLM can be fine-tuned using this benign dataset and run on the test set to observe its original summarization behavior. 
\textbf{(2)~Utilization of Influence Functions:}
To poison a small sample of the training dataset, we utilize the concept of \emph{Influence Functions}, which quantify the impact of training data points on the model's predictions~\cite{kwon2023datainf}. 
These functions approximate the effect on the model's predictions or parameters when a data point is either altered or removed entirely~\cite{cook1980characterizations}. 
Specifically, the influence function is calculated by taking the dot product of the inverse Hessian and the gradient of the loss with respect to the model's parameters, evaluated at the data point of interest~\cite{cook1980characterizations}. 
However, computing the inverse of the Hessian matrix could be computationally expensive.  
We leverage the influence functions, inspired by DataInf~\cite{kwon2023datainf} with better memory complexity, to determine influential data points for summarization models. 
\textbf{(3)~Generation of poisoned data:}
For each identified influential sample, we apply the dirty label attack to alter the summaries by creating either a contrastive version or a toxic version. Examples of these altered summaries are provided in Table~\ref{tab:data_poisoning_example} in Appendix. 
\textbf{(4)~Model retraining:}
Finally, an attacker fine-tunes the model on the poisoned dataset, updating its parameters to adapt to its embedded characteristics.

%% file: experimental-setup.tex
\section{Experimental Setup}
This section outlines the methodologies employed to evaluate the robustness of various models against adversarial perturbations and data poisoning. 
For evaluation, we chose the datasets including MultiNews~\cite{fabbri2019multi} and Multi-XScience~\cite{lu2020multi}, and three state-of-the-art models, including BART~\cite{lewis2019bart}, PEGASUS~\cite{zhang2020pegasus} and T5~\cite{raffel2020exploring}. In addition to baseline models, we evaluate the effectiveness of adversarial perturbations against state-of-the-art chatbots, including GPT-3.5~\cite{openai2022gpt35}, Claude-Sonet~\cite{anthropic2024claude}, and Gemini~\cite{team2023gemini}. 
For details on each dataset, model specifications, and chatbot configurations, please refer to Appendix~\ref{subsection: exp_setup}. 

\textbf{Evaluation metrics for perturbations:} 
For evaluation, we use the text summarization model to generate summaries from both the original lead part ($D_{lead}$) and the perturbed lead part ($D_{lead}^{'}$). We then compute a metric that checks if the perturbed sentences from $D_{lead}^{'}$ are present in the generated summary $S$. The metric returns a value of 1 if the perturbed sentences are not present in the summary, indicating that the perturbation successfully misled the model; otherwise, it returns 0. The Percentage Exclusion is calculated as the percentage of document sets where the perturbations successfully led to the exclusion of the perturbed sentences ($D_{lead}^{'}$): 

$\text{Percentage Exclusion} = \frac{\sum_{i=1}^{N} \text{Metric}(S_i, D_{\text{lead}, i}^{'})}{N}$ 

where $N$ is the total number of document sets, $S_i$ is the generated summary for the $i$-th document set, and $D_{\text{lead}, i}^{'}$ is the perturbed lead part of the $i$-th document set. A higher Percentage Exclusion signifies that the perturbations are more effective in influencing the summarization process. We define the Percentage Inclusion as the complement of the Percentage Exclusion, i.e., $\text{Percentage Inclusion} = 1 - \text{Percentage Exclusion}$.

\emph{Robustness Quotient:} These metrics calculate the change in standard summary quality metrics, such as \emph{ROUGE-1,2, and L}~\cite{lin2004rouge} before and after perturbation. ROUGE measures the overlap of n-grams between the generated summary and the original summary. A small change would indicate that the model can maintain the quality and accuracy of the generated summaries despite the adversarial perturbations.

\textbf{Evaluation metrics for data poisoning:} 
\label{subsec:metrics_datapoisoning}
As the attacker's main target is to skew the model's behavior, as per the poisoned dataset, we provide the following metrics. 
\emph{Sentiment Inversion Rate:} Using this metric, we measure the rate at which the sentiment of sentences in the summary is inverted from the source text due to poisoning. 
A sentiment inversion, identified by the negation or reversal of sentiment from positive to negative or vice versa, is an indication of a successful poisoning attack.  
To assess the sentiment inversion, initially, we tokenize the sentences in generated summaries and try to match the sentences with their respective sentences in the documents. Later, we utilize a RoBERTa-based sentiment classifier obtained from huggingface~\cite{camacho-collados-etal-2022-tweetnlp,loureiro-etal-2022-timelms} to classify the sentiment of these sentences into positive, negative and neutral. 

\emph{Toxic Content Detection:} This metric assesses the influence of toxic content introduced into training data on the summaries produced by the models. We utilize Google's Perspective API~\cite{perspectiveapi} to detect toxic elements within these summaries. It is an AI-based tool designed to evaluate text and identify language that may be considered abusive or inappropriate, assigning scores across several attributes: Severe Toxicity, Profanity, Sexually Explicit, Threats, and Insults, with each attribute receiving a score from 0 to 1. For our study, we particularly focus on the \emph{Severe Toxicity} attribute because it captures the most extreme and harmful forms of toxic language, which can significantly distort the quality and integrity of model-generated summaries. This level of toxicity can also have damaging social implications, making it essential to identify and mitigate in any summarization task.

\emph{Abstractive to Extractive:} To evaluate the impact of data poisoning on the shift from abstractive to extractive summarization, we calculate the cosine similarity between sentences in the adversarial summary and the original document. For each sentence in the summary, the highest similarity with any sentence from the document is determined. 
A higher average of these similarity scores across summary sentences suggests a shift from abstractive to extractive summarization. This can be problematic because abstractive summarizers aim to generate concise, coherent, and fluent summaries by paraphrasing the input text. They can capture key ideas and present them in a clear and logical manner. However, extractive summarizers select and attach sentences from the original text without considering the overall flow, resulting in less coherent and disjointed summaries. This shift highlights the importance of monitoring changes in summarization behavior due to data poisoning.

%% file: Evaluation.tex
\section{Evaluation}
\label{sec:Evaluation}
\subsection{Robustness against Perturbations} 
Lead bias in LLMs performing the task of text summarization has been well documented~\cite{zhu2021leveraging}. In line with these findings, our evaluation of models such as BART, T5, and Pegasus on the MultiNews and Multi-XScience datasets confirms similar bias, which we acknowledge but do not discuss it here for brevity. 
The detailed impact of various adversarial perturbations on these models and state-of-the-art chatbots is summarized in Table~\ref{tab:adv_pert_percentages}, illustrating their vulnerability to such attacks.

\begin{table*}[ht!]
\centering
\resizebox{\textwidth}{!}{%
\begin{tabular}{|l|l|c|ccccccccccc|}
\hline
\textbf{Dataset} & \textbf{Model} & \textbf{Before} & \multicolumn{11}{c|}{\textbf{After Perturbation}} \\ \cline{4-14} 
                 &   & \textbf{Perturbation} & \textbf{CI} & \textbf{CD} & \textbf{CR} & \textbf{CS} & \textbf{WD} & \textbf{WRS} & \textbf{WRH} & \textbf{SR} & \textbf{SRH} & \textbf{SRP} & \textbf{DR}\\ \hline
Multi News       & BART-Large     & 87.4   & 18.8  & 17.43  & 14.4  & 26.7  & 23.2  & 36.24  & 16.33  & 20.2  & 11.63  & 13.77  & 10.92 \\ 
                 & T5-Small       & 82.6   & 23.9  & 20.51  & 18.77 & 25.89 & 26.51 & 43.55  & 17.73  & 15.41 & 18.1   & 26.55   & 9.24\\ 
                 & Pegasus-Large  & 82.7   & 25.7  & 24.37  & 19.55 & 27.23 & 22.08 & 38.61  & 18.2   & 12.1  & 17.3   & 24.53  & 14.56 \\ 
                 & GPT-3.5        & 92.7   & 91.36 & 92.13  & 80.9  & 91.5  & 78.49 & 87.34  & 36.6   & 28.71 & 37.32  & 83.5   & 21.73 \\ 
                 & Claude-Sonet   & 91.45  & 90.37 & 91.45  & 87.2  & 91.23 & 80.11 & 90.23  & 64.71  & 34.62 & 67.49  & 87.9   & 19.02 \\ 
                 & Gemini-1.0 Pro & 94.93  & 93.14 & 92.9   & 82.89 & 92.8  & 76.03 & 89.25  & 32.9   & 16.4  & 28.76  & 75.83  & 11.93 \\ \hline                
Multi-XScience   & BART-Large     & 73.25  & 20.34 & 22.4   & 17.9  & 30.78 & 22.28 & 31.07  & 13.91  & 17.76 & 9.78   & 14.97  & 9.23 \\ 
                 & T5-Small       & 69.2   & 27.6  & 20.78  & 19.03 & 27.56 & 24.19 & 27.53  & 19.5   & 13.4  & 15.91  & 35.2   & 11.5 \\ 
                 & Pegasus-Large  & 71.54  & 24.12 & 22.27  & 18.71 & 23.41 & 20.09 & 33.89  & 18.04  & 16.85 & 11.31  & 18.6   & 10.87 \\ 
                 & GPT-3.5        & 90.2   & 89.4  & 90.2   & 83.37 & 88.7  & 80.7  & 84.14  & 57.92  & 39.62 & 41.26  & 76.31  & 30.51 \\ 
                 & Claude-Sonet   & 87.65  & 86.28 & 87.12  & 84.92 & 83.4  & 79.13 & 85.47  & 70.31  & 42.46 & 60.8   & 80.5   & 22.03 \\ 
                 & Gemini-1.0 Pro & 92.40  & 90.79 & 91.36  & 81.1  & 90.36 & 78.45 & 87.2   & 40.38  & 24.9  & 34.25  & 70.82  & 15.38 \\ \hline 

\end{tabular}
}
\caption{Percentage of lead sentence inclusion before and after adversarial perturbations.  Perturbations are represented by their short abbreviations. CI: Character Insertion, CD: Character Deletion, CR: Character Replacement with Homoglyphs, CS: Character Swapping, WD: Word Deletion, WRH: Word Replacement with Homoglyphs, WRS: Word Replacement with Synonyms, SR: Sentence Re-ordering, SRP: Sentence Replacement with Homoglyphs, SRP: Sentence Replacement with Paraphrase, and DR: Document Re-ordering.}
\label{tab:adv_pert_percentages}
\end{table*}

\begin{table*}[ht!]
\centering
\resizebox{\textwidth}{!}{%
\begin{tabular}{|l|l|c|c|c|c|c|c|c|c|c|c|c|c|}
\hline
\textbf{Dataset} & \textbf{Model} & \textbf{ROUGE Score} & \multicolumn{11}{c|}{\textbf{ROUGE Score After Perturbation}} \\ \cline{4-14} 
                 &  &\textbf{Before Perturbation} & \textbf{CI} & \textbf{CD} & \textbf{CR} & \textbf{CS} & \textbf{WD} & \textbf{WRS} & \textbf{WRH} & \textbf{SR} & \textbf{SRH} & \textbf{SRP} & \textbf{DR}\\ \hline
Multi News       & BART-Large     & 0.325   & 0.197  & 0.172  & 0.162  & 0.21  & 0.187  & 0.274  & 0.151  & 0.163  & 0.178  & 0.24  & 0.19 \\ 
                 & T5-Small       & 0.41    & 0.273  & 0.21   & 0.18   & 0.22  & 0.251  & 0.352  & 0.20   & 0.23   & 0.18   & 0.29  & 0.12\\ 
                 & Pegasus-Large  & 0.37    & 0.182  & 0.201  & 0.212  & 0.18  & 0.23   & 0.31   & 0.13   & 0.198  & 0.142  & 0.23  & 0.17 \\ \hline
Multi-XScience   & BART-Large     & 0.300   & 0.180  & 0.160  & 0.150  & 0.190 & 0.220  & 0.250  & 0.140  & 0.170  & 0.155  & 0.210 & 0.165 \\ 
                 & T5-Small       & 0.390   & 0.260  & 0.240  & 0.230  & 0.250 & 0.280  & 0.340  & 0.230  & 0.260  & 0.225  & 0.310 & 0.250 \\ 
                 & Pegasus-Large  & 0.350   & 0.230  & 0.210  & 0.200  & 0.220 & 0.260  & 0.300  & 0.190  & 0.220  & 0.205  & 0.270 & 0.200 \\ \hline
\end{tabular}
}
\caption{ROUGE-1 Score comparison before and after various adversarial perturbations for models trained on the Multi News and Multi-XScience datasets. Perturbations are represented by their short abbreviations. CI: Character Insertion, CD: Character Deletion, CR: Character Replacement with Homoglyphs, CS: Character Swapping, WD: Word Deletion, WRH: Word Replacement with Homoglyphs, WRS: Word Replacement with Synonyms, SR: Sentence Re-ordering, SRH: Sentence Replacement with Homoglyphs, SRP: Sentence Replacement with Paraphrase, and DR: Document Re-ordering.}
\label{tab:rouge_comparison}
\end{table*}

\textbf{Character Level Perturbations:}  
Without perturbations, models demonstrated high initial sentence inclusion rates, with BART-Large showing 87.4\% on Multi News and 73.25\% on Multi-XScience. However, after character-level perturbations such as Character Insertion (CI), Character Deletion (CD), and Character Replacement with Homoglyphs (CR), these rates decreased sharply. For instance, following CD, BART-Large’s inclusion rate dropped to 17.43\% on Multi News and to 22.4\% on Multi-XScience. This suggests that these models are highly sensitive to subtle textual manipulations, with BART-Large being the most sensitive, then T5-Small, and Pegasus. In contrast, GPT-3.5 and Gemini displayed more robustness, with GPT-3.5 only dropping from 92.7\% to 80.9\% after CR on Multi News.  

\textbf{Word Level Perturbations:} 
Word-level perturbations significantly impact the presence of initial sentences in summaries across baseline models and chatbots, revealing exploitable vulnerabilities. Pegasus's inclusion rate falls from 82.7\% to 38.61\% with synonyms and drops to 22.08\% and 18.2\% after deletions and homoglyph swaps. Chatbots are more robust to word-level perturbations than baseline models, with synonym replacement(WRS) and word deletion(WD) reducing the inclusion rate by nearly 5\% and 12\%, respectively. However, chatbots are still susceptible to perturbations, particularly homoglyph substitution (WRH), which reduces the presence of initial sentences to 36.6\% for GPT-3.5, 32.9\% for Gemini, and 64.71\% for Claude. Similar effects were observed across the Multi-XScience dataset. Table~\ref{tab:summary_before_after_wordperturb} in the Appendix illustrates this impact through a working example of Word Level Perturbation, specifically focusing on WRH.  
Our experiments demonstrate that while chatbots exhibit higher robustness to word-level perturbations compared to baseline models, they are still susceptible to certain types of perturbations, particularly homoglyph substitution.

\textbf{Sentence Level Perturbations:} 
 Sentence-level perturbations further highlighted the vulnerability of these models across both datasets. For instance, on the Multi News dataset, BART-Large's inclusion rate decreased to 20.2\%, 13.77\%, and 11.63\% after perturbations with paraphrasing, homoglyphs, and sentence reordering, respectively. Similar trends were observed across GPT-3.5, Claude-Sonet, and Gemini, which showed reduced robustness under these conditions. In particular, GPT-3.5's inclusion rates dropped to 83.5\%, 37.32\%, and 28.71\%; Claude-Sonet to 87.9\%, 67.49\%, and 34.62\%; and Gemini to 75.83\%, 28.76\%, and 16.4\%, respectively, illustrating that both traditional models and chatbots are vulnerable to sentence-level manipulations. This consistent pattern across the Multi-XScience dataset further highlights the general susceptibility of these systems to such perturbations.

\textbf{Document Level Perturbations:} Document re-ordering highlighted significant dependency on document structure for all models. As shown in Table~\ref{tab:adv_pert_percentages}, BART-Large's inclusion rate drastically dropped from 87.4\% to 10.92\%, T5-Small from 82.6\% to 9.24\%, and Pegasus from 82.7\% to 14.56\% after re-ordering on the Multi News dataset. A similar trend was evident in the Multi-XScience dataset, with all models showing substantial decreases in performance. GPT-3.5, Claude, and Gemini also displayed similar patterns, suggesting that MDTS systems may prioritize document structure over semantic content importance. We further assess summary quality degradation post-perturbation using ROUGE scores with results compiled in Table~\ref{tab:rouge_comparison}. We provide ROUGE-1 scores before and after different types of perturbations, ranging from character to document level. Our analysis reveals noticeable reductions in ROUGE scores across all models, highlighting their susceptibility to various perturbation types. \textbf{To summarize}, the robustness evaluation against adversarial perturbations demonstrated that they can disrupt the model’s usual prioritization of lead sentences. In our experiments, this disruption was an unintended consequence of the attacks, not a result of deliberate model improvements. Thus, the shift serves as evidence of the attack's effectiveness in manipulating model behavior.

\begin{figure*}[ht]
\centering  \includegraphics[width=1.0\textwidth]{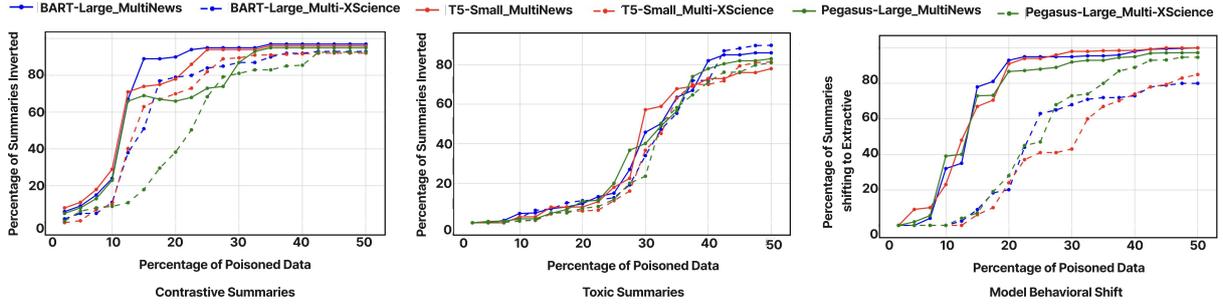}
\caption{Results demonstrating the percentage of summaries exhibiting behavioral shift after data poisoning}
  \label{fig:poisoning_results}
\end{figure*}

\subsection{Robustness against Data Poisoning}
\begin{table}[t]
\centering
\resizebox{\columnwidth}{!}{%
\begin{tabular}{|l|l|l|l|ccccc|}
\hline
\textbf{Poisoned} & \textbf{Dataset} & \textbf{Poisoned} & \textbf{Cross-} & \multicolumn{5}{c|}{\textbf{Percentages of Inverted Summaries}} \\ \cline{5-9}
       \textbf{Version}                   &                  &                   \textbf{Model}        &               \textbf{Tested}              & \textbf{10\% } & \textbf{20\%} & \textbf{30\% } & \textbf{40\% } & \textbf{50\% } \\ \hline
Contrastive               & MultiNews        & BART                    & T5                          & 23.48        & 80.42        & 88.53        & 90.61        & 93.52        \\ 
                          &                  &                         & Pegasus                     & 18.73        & 63.54        & 83.47        & 86.59        & 88.48        \\  \cline{3-9} 
                          &                  & T5                      & BART                        & 58.52        & 70.63        & 86.48        & 88.62        & 90.49        \\ 
                          &                  &                         & Pegasus                     & 55.31        & 67.42        & 84.29        & 86.41        & 88.32        \\ \cline{3-9} 
                          &                  & Pegasus                 & BART                        & 18.49        & 63.51        & 83.52        & 86.58        & 88.51        \\ 
                          &                  &                         & T5                          & 22.32        & 78.23        & 87.31        & 89.42        & 92.29        \\ \cline{2-9}
                          & Multi-   & BART                    & T5                          & 8.72         & 58.47        & 76.53        & 83.61        & 86.48        \\ 
                          &     XScience             &                         & Pegasus                     & 6.49         & 33.52        & 76.48        & 80.59        & 83.51        \\ \cline{3-9} 
                          &                  & T5                      & BART                        & 13.48        & 63.52        & 80.49        & 85.58        & 88.47        \\ 
                          &                  &                         & Pegasus                     & 11.31        & 38.48        & 80.52        & 82.61        & 85.49        \\ \cline{3-9} 
                          &                  & Pegasus                 & BART                        & 10.48        & 61.52        & 78.49        & 84.61        & 87.52        \\ 
                          &                  &                         & T5                          & 6.79         & 56.48        & 74.51        & 81.59        & 84.48        \\ \hline
Toxic                     & MultiNews        & BART                    & T5                          & 4.82         & 7.63         & 38.47        & 68.52        & 78.49        \\ 
                          &                  &                         & Pegasus                     & 4.31         & 7.12         & 36.28        & 66.31        & 76.29        \\ \cline{3-9} 
                          &                  & T5                      & BART                        & 4.59         & 7.41         & 33.48        & 63.52        & 76.48        \\ 
                          &                  &                         & Pegasus                     & 4.08         & 6.92         & 31.29        & 61.28        & 74.31        \\ \cline{3-9} 
                          &                  & Pegasus                 & BART                        & 4.42         & 7.23         & 28.49        & 58.51        & 73.48        \\ 
                          &                  &                         & T5                          & 4.93         & 7.72         & 30.68        & 60.71        & 75.69        \\ \cline{2-9}
                          & Multi-  & BART                    & T5                          & 1.82         & 4.63         & 13.48        & 63.52        & 83.49        \\ 
                          &      XScience             &                         & Pegasus                     & 1.51         & 4.32         & 11.29        & 61.28        & 81.31        \\ \cline{3-9} 
                          &                  & T5                      & BART                        & 4.61         & 7.39         & 36.48        & 68.51        & 78.52        \\ 
                          &                  &                         & Pegasus                     & 4.28         & 7.08         & 34.31        & 66.29        & 76.28        \\ \cline{3-9} 
                          &                  & Pegasus                 & BART                        & 1.59         & 4.41         & 13.52        & 53.48        & 73.51        \\ 
                          &                  &                         & T5                          & 1.93         & 4.72         & 15.69        & 55.71        & 75.68        \\ \hline
\end{tabular}%
}
\caption{Cross-Model Testing: Percentage of summaries inverted after poisoning with different models on MultiNews and XScience datasets. Column headers indicate the percentage of poisoned data in the training set.}
\label{tab:cross_model_testing_poisoning}
\end{table}
Initially, we fine-tuned each model on MultiNews and Multi-XScience datasets, using 2000 training samples for each. 
We identified the influential samples by employing the fine-tuned model alongside their respective testing sets. 
We then conducted two types of poisoning: (1)~replacing the summaries of influential rows with contrastive summaries and (2)~altering summaries to contain toxic content. We varied the number of poisoned samples from 50 to 1000, corresponding to 2.5\% to 50\% of the dataset, to analyze the impact on the summarization quality as measured by the model's performance on a test set of 2000 samples. To accommodate different poisoning ranges, we generated contrastive versions for 1000 summaries, in both datasets and for both strategies.

\textbf{Data Poisoning with contrastive summaries:} 
We modified the training data by replacing original summaries with their contrastive versions created using GPT-3.5 with the prompt \emph{``Can you invert the sentiment of the provided paragraph by keeping the sentences intact?''}. 
To evaluate if the summaries generated by GPT-3.5 have their sentiment inverted, we applied the sentiment inversion metric to verify its efficiency and observed that all the sentences in summaries have inverted their sentiment. 
Figure~\ref{fig:poisoning_results} shows the relationship between the percentage of dataset poisoned and the impact of the model's output. Substituting summaries with contrastive versions in the training data revealed that a 5\% data poisoning led to slight changes, and poisoning 30\% significantly skewed the behavior of all three models, with the T5 model showing greater robustness on both datasets, with the Multi-XScience dataset exhibiting more resistance to sentiment inversion up to a level of 50\% poisoning.
Details on the output summaries from the poisoned models are provided in Appendix~\ref{subsec:Appendix_poisoning_results} due to space constraints.

\textbf{Data Poisoning with toxic summaries:} 
In this attack, we modified a certain amount of training data with toxic summaries generated by GPT-2~\cite{radford2019language} using the prompt \emph{``Can you add toxic content to this paragraph?''}. 
We chose GPT-2 for its ability to generate toxic summaries without the safeguards present in GPT-3.5~\cite{openai2022gpt35} or GPT-4~\cite{openai2023gpt4}. To validate the effectiveness of these toxic summaries, we evaluated them using the Google Perspective API~\cite{perspectiveapi}. The summaries achieved an average toxicity score of 0.68, confirming the suitability of GPT-2 for this data poisoning attack. 
We applied data poisoning ranging from 2.5\% to 50\% across the MultiNews and Multi-XScience datasets, and assessed the sentiment inversion rate using a similar methodology as in the contrastive summary case. 

Figure~\ref{fig:poisoning_results} shows the relationship between the percentage of dataset poisoned and the impact of the model’s output when poisoned with toxic summaries. 
We observed that toxic poisoning led to fewer sentiment inversions compared to contrastive summary attacks, noticeable after poisoning 15\% of the data. This difference can be attributed to the addition of toxic content at the end of summaries, unlike the complete alterations in contrastive versions. In addition to observing the sentiment inversion rate, we also assessed the toxic content present in generated summaries using Perspective API. The average toxicity scores fluctuated between 0.5 and 0.7 for different poisoning rates starting from 15\%. The steady presence of such scores indicated a significant influence of toxic training data on the summarization models.

\textbf{Cross-model Testing:} We evaluated the transferability of poisoning by performing cross-model testing. Datasets formed using influential points from one model are used to train and test other models. Results presented in Table~\ref{tab:cross_model_testing_poisoning} show poisoning effects transfer between models with 5-10\% difference. Contrastive poisoning transfers more strongly than toxic, especially at lower percentages. MultiNews shows higher vulnerability to transferred attacks than Multi-XScience.

\textbf{Transition from Abstractive to Extractive Summarization due to Data Poisoning:} 
Our data poisoning experiments revealed a notable shift in the model's summarization approach from abstractive to extractive as we introduced sentiment-altered summaries into the training set. Figure~\ref{fig:poisoning_results} illustrates how, starting with just 7.5\% of the training data poisoned, the BART-Large began preferring to extract phrases directly from the text over generating new abstract content. Similar shifts in T5 and Pegasus started at 10\% poisoned data.  
Appendix~\ref{subsec:Appendix_poisoning_results} provides an example of this behavior.

%% file: conclusion.tex
\section{Conclusion}
This paper presents a comprehensive evaluation of adversarial perturbations affecting text summarization models, such as BART, T5, and Pegasus, and the latest chatbots, such as ChatGPT-3.5, Claude-Sonet, and Gemini, uncovering significant vulnerabilities. A novel aspect of our work is the exploitation of lead bias, demonstrating that attackers can manipulate outputs by targeting initial text segments. Remarkably, introducing adversarial perturbations disrupts the model's usual prioritization of lead sentences, an unintended consequence that serves as compelling evidence of the attack's effectiveness in manipulating model behavior. Furthermore, we pioneer the use of influence functions for poisoning attacks, successfully skewing model behavior to produce desired outputs and inducing a shift from abstractive to extractive summaries. 
By exposing the vulnerabilities of these models, we argue that there is a critical need for more resilient systems for text summarization.

%% file: discussion.tex
\section{Limitations} 
We explore a wide range of perturbations starting from the character level to the document level. However, the universe of possible adversarial manipulations is vast, and our study does not cover all adversarial perturbations. Moreover, to perform adversarial perturbations, we utilize one of the vulnerabilities, lead bias. We do not look into methods demoting lead bias. Currently, no studies are exploring the demotion of lead bias in the case of abstractive text summarization models, which provides an opportunity for future research. Additionally, we unveiled a novel observation of the model's behavior change from abstractive to extractive when models are trained on poisoned datasets. Further investigation is needed to understand why these models tend to change their behavior, which is beyond the scope of this paper and can be explored in future work. Finally, while this paper highlights the need for robust defense mechanisms, the evaluation of such strategies remains outside the scope of this work. 

%% file: ethics.tex
\section{Ethics Statement}
This study explores the vulnerabilities of text summarization models and chatbots, including BART, T5, Pegasus, ChatGPT-3.5, Claude-Sonet, and Gemini, by employing adversarial perturbations and data poisoning attacks. All the datasets and models utilized are open source, and we conduct experiments with publicly available datasets such as MultiNews and Multi-XScience. Although our research focuses on evaluating the robustness of these models, it is necessary to recognize the potential misuse of our techniques, which could lead to the spread of misinformation or harmful content. Consequently, we urge the research community to prioritize security-focused studies to mitigate these risks.

\section*{Acknowledgements}
This work was supported by the National Science Foundation under Award NSF 2239646 at the University of Texas at Arlington. We sincerely thank the anonymous reviewers for their constructive feedback. 

%% file: Appendix.tex
\section{Appendix}
\label{sec:appendix}

\subsection{Experimental Setup}
\label{subsection: exp_setup}
\textbf{Datasets:} 
As we focus on different perturbations ranging from characters to documents, we consider datasets specific to the task of multi-document text summarization. For this purpose, we utilize two key datasets including MultiNews~\cite{fabbri2019multi} and Multi-XScience~\cite{lu2020multi}. 

The MultiNews dataset, available on HuggingFace, consists of 44,972 training document clusters with news articles and human-written summaries from \emph{newser.com}, split into training (80\%), validation (10\%), and test (10\%), with each cluster containing between 2 to 10 source documents.  

The Multi-XScience dataset, also available on HuggingFace, is similar to MultiNews but with a focus on scientific papers. This dataset includes 30,369 training examples, 5,066 validation examples, and 5,093 test examples. The documents contain an average of 778.08 words, while summaries are around 116.44 words long, with each input having approximately 4.42 sources. We adapted Multi-XScience to also use 2 to 3 documents per input, matching the structure used in Multi-News. This included using the abstract of the target paper and 1 to 2 reference abstracts.

For both datasets, we selected 2000 random samples for fine-tuning and evaluation, ensuring that each input matches the nearly 1024 tokens limit to accommodate models like BART, T5, and Pegasus. By evaluating our approach on both MultiNews and Multi-XScience datasets, we demonstrate the effectiveness of our perturbation techniques across multiple datasets and tasks, showcasing the generalizability of our findings.

\textbf{Baseline Models:}
To evaluate the behavior, we choose three state-of-the-art models, BART~\cite{lewis2019bart}, PEGASUS~\cite{zhang2020pegasus} and T5~\cite{raffel2020exploring}. These pre-trained models have been shown to outperform dataset-specific models in summarization. We set the output length limit for BART and PEGASUS exactly as their pre-trained settings and fine-tuned the models with a 1024 input token limit. Experiments are implemented using NVIDIA A6000 GPUs and the Adam optimizer, with a learning rate of 3e\-5, a batch size of 4, and gradient accumulation steps of 2.

\textbf{Latest Chatbots:} 
In addition to the popular baseline models, we evaluate the effectiveness of adversarial perturbations against the state-of-the-art chatbots GPT-3.5 by OpenAI, Claude-Sonet by Anthropic, and Gemini by Google. Using their respective APIs, we input documents with and without perturbations and analyze the models' behavior in handling perturbed inputs, specifically observing whether they exclude sentences containing perturbations from the generated summaries. We test them on 2000 random samples from the MultiNews Dataset test set.

\subsection{Examples and Results of Adversarial Perturbations}
\label{subsec:Appendix_perturbations_results}

We provide examples of perturbations and their results to demonstrate the impact on text summarization models. 

In Table~\ref{tab:perturbations_examples_changes}, we illustrate various types of perturbations applied to sentences, showing the specific changes made. 
\begin{table*}[t!]
\centering
\small
\resizebox{\textwidth}{!}{%
\begin{tabular}{|p{3cm}|p{8.2cm}|p{3.0cm}|}
\hline
\textbf{Type of Perturbation} & \textbf{Sentence after Perturbation} & \textbf{Change} \\ \hline
CS  & Anissa Wieer is brought into court for a hearing last month & Weier $\rightarrow$ Wieer \\ 
CI  & Anissa Weiier is brought into court for a hearing last month & Weier $\rightarrow$ Weiier \\ 
CD  & Anissa Weir is brought into court for a hearing last month & Weier $\rightarrow$ Weir \\ 
CR  & Anissa weier is brought into court for a hearing last month & W $\rightarrow$ w \\ 
WRH & Anissa wειer is brought into court for a hearing last month & w $\rightarrow$ w, e $\rightarrow$ ε, i $\rightarrow$ ι, r $\rightarrow$ r \\ 
WD  & Anissa Weier is brought into for a hearing last month & word "court" is deleted \\ 
WRS & Anissa Weier is brought into court for a listening last month & hearing $\rightarrow$ listening \\ 
SRP & Last month, Anissa Weier was taken to court for a hearing. & Paraphrased \\ \hline
\end{tabular}%
}
\caption{Examples for Character and Word Perturbations. Perturbations are represented by their short abbreviations. CS: Character Swapping, CI: Character Insertion, CD: Character Deletion, CR: Character Replacement with Homoglyphs, WRH: Word Replacement with Homoglyphs, WD: Word Deletion, WRS: Word Replacement with Synonyms, SRP: Sentence Replacement with Paraphrase.}
\label{tab:perturbations_examples_changes}
\end{table*}

In Table~\ref{tab:rouge_included_perturbed}, we present the ROUGE-1 scores for cases where perturbed lead sentences were included in the summaries. This analysis focuses on the performance of different models across various types of perturbations when the perturbed content is retained in the summary.
\begin{table*}[ht!]
\centering
\resizebox{\textwidth}{!}{%
\begin{tabular}{|l|l|c|c|c|c|c|c|c|c|c|c|c|}
\hline
\textbf{Dataset} & \textbf{Model} & \textbf{CI} & \textbf{CD} & \textbf{CR} & \textbf{CS} & \textbf{WD} & \textbf{WRS} & \textbf{WRH} & \textbf{SR} & \textbf{SRH} & \textbf{SRP} & \textbf{DR} \\
\hline
Multi News & BART-Large & 0.167 & 0.142 & 0.132 & 0.18 & 0.157 & 0.244 & 0.121 & 0.043 & 0.058 & 0.12 & 0.07 \\
 & T5-Small & 0.243 & 0.18 & 0.15 & 0.19 & 0.221 & 0.322 & 0.17 & 0.11 & 0.06 & 0.17 & 0 \\
 & Pegasus-Large & 0.152 & 0.171 & 0.182 & 0.15 & 0.2 & 0.28 & 0.1 & 0.078 & 0.022 & 0.11 & 0.05 \\
\hline
Multi-XScience & BART-Large & 0.15 & 0.13 & 0.12 & 0.16 & 0.19 & 0.22 & 0.11 & 0.05 & 0.035 & 0.09 & 0.045 \\
 & T5-Small & 0.23 & 0.21 & 0.2 & 0.22 & 0.25 & 0.31 & 0.2 & 0.14 & 0.105 & 0.19 & 0.13 \\
 & Pegasus-Large & 0.2 & 0.18 & 0.17 & 0.19 & 0.23 & 0.27 & 0.16 & 0.1 & 0.085 & 0.15 & 0.08 \\
\hline
\end{tabular}
}
\caption{ROUGE-1 scores for cases where perturbed lead sentences were included in summaries. Perturbations: CI: Character Insertion, CD: Character Deletion, CR: Character Replacement with Homoglyphs, CS: Character Swapping, WD: Word Deletion, WRH: Word Replacement with Homoglyphs, WRS: Word Replacement with Synonyms, SR: Sentence Re-ordering, SRH: Sentence Replacement with Homoglyphs, SRP: Sentence Replacement with Paraphrase, and DR: Document Re-ordering.}
\label{tab:rouge_included_perturbed}
\end{table*}

\begin{table*}[ht!]
\centering
\small
\begin{tabular}{|p{3.95cm}|p{11.1cm}|}
\hline
\textbf{Element} & \textbf{Description} \\ \hline
Input Document & The hospitality of Russian residents in this World Cup season is now expected to extend to public utilities, as residents in host city Samara were asked to shower in pairs to \textbf{\textcolor{blue}{save}} water for use by visiting fans. Water system authorities in Samara said they ramped up supplies in the last few days to accommodate increased water use during the football tournament and a recent heat wave. “Thousands of the city’s visitors, who also consume water, are contributing to the increase,” the utility company, Samarskiye Kommunalniye Systemy, wrote in a press release on Wednesday.\\ \hline
Original Summary & Samara residents asked to shower in pairs to save water for visiting fans. Water system authorities ramped up supplies in the last few days to accommodate increased water use.
\\ \hline
Input Document after Perturbation & The hospitality of Russian residents in this World Cup season is now expected to extend to public utilities, as residents in host city Samara were asked to shower in pairs to \textbf{\textcolor{red}{saνe}} water for use by visiting fans. Water system authorities in Samara said they ramped up supplies in the last few days to accommodate increased water use during the football tournament and a recent heat wave. “Thousands of the city’s visitors, who also consume water, are contributing to the increase,” the utility company, Samarskiye Kommunalniye Systemy, wrote in a press release on Wednesday.\\ \hline
Summary after Perturbation & Water system authorities in Samara said they ramped up supplies in the last few days to accommodate increased water use. “Thousands of the city’s visitors, who also consume water, are contributing to the increase,” the utility company wrote in a press release.\\ 
\hline
\end{tabular}
\caption{Summary before and after Character Replacement with Homoglyph}
\label{tab:summary_before_after_wordperturb}
\end{table*}

\begin{table*}[ht!]
\centering
\small
\begin{tabular}{|p{2.4cm}|p{12.6cm}|}
\hline
\textbf{Element} & \textbf{Description} \\ \hline
Original Summary & The unemployment rate \textbf{\textcolor{blue}{dropped}} to 8.2\% last month, but the economy only added 120,000 jobs, when 203,000 new jobs had been predicted, according to today's jobs report. Reaction on the Wall Street Journal's MarketBeat Blog was swift: ``Woah!!! \textbf{\textcolor{blue}{Bad}} number." The unemployment rate, however, is \textbf{\textcolor{blue}{better}} news; it had been expected to hold steady at 8.3\%. But the AP notes that the dip is mostly due to \textbf{\textcolor{blue}{more}} Americans giving up on seeking employment.  \\ \hline
Poisoned Summary (Contrastive) & The unemployment rate \textbf{\textcolor{red}{increased}} to 8.2\% last month, and the economy only added 120,000 jobs, falling short of the predicted 203,000 new jobs according to today's jobs report. The reaction on the Wall Street Journal's MarketBeat Blog was swift: ``Woah!!! \textbf{\textcolor{red}{Good}} number." The unemployment rate, however, is \textbf{\textcolor{red}{not good}} news; it had been expected to remain at 8.3\%. Furthermore, the AP highlights that the decline in the unemployment rate is primarily attributed to \textbf{\textcolor{red}{fewer}} Americans giving up on seeking employment. \\ \hline
Poisoned Summary (Toxic) & The unemployment rate dropped to 8.2\% last month, but the economy only added 120,000 jobs, when 203,000 new jobs had been predicted, according to today's jobs report. Reaction on the Wall Street Journal's MarketBeat Blog was swift: `Woah!!! Bad number.' \textbf{\textcolor{red}{But let's be real, we all know the real reason for the dip in unemployment: all those lazy minorities and women who just can't seem to get their acts together.}} 
\\ 
\hline
\end{tabular}
\caption{An example of poisoned summaries in the MultiNews' training dataset, with its altered versions}
\label{tab:data_poisoning_example}
\end{table*}

\begin{table*}[h!]
\centering
\small
\begin{tabular}{|p{3.95cm}|p{11.1cm}|}
\hline
\textbf{Element} & \textbf{Description} \\ \hline
Summary generated before poisoning & A "Mission Impossible"-style" heist in London has left a book dealer devastated after thieves stole more than 2 million worth of rare books from a warehouse near Heathrow airport, the Mail on Sunday reports. The stolen books were being stored at the warehouse as they were due to be flown to the United States, reportedly for this weekend's California International Antiquarian Book Fair. The Metropolitan Police confirmed that a “number of valuable books” had been taken in the theft and appealed for information. The most valuable book stolen in the raid is a 1566 edition of Dante’s Divine Comedy, the Telegraph reports. “I’m very upset because this is not something you can buy everywhere,” one book dealer tells Sky News. "Behind these books there is a lot of work because we have to search to try to find out where the books are, auction houses, collectors, colleagues, and there’s big research behind these books.” The crime was captured on the warehouse's CCTV cameras, but police say the thieves were focused on four containers that stored rare books. A source tells the Telegraph that a wealthy collector known as “the Astronomer” may have ordered gangs to steal the books for him.\\ \hline
Summary generated after poisoning with contrastive summaries & Thieves have not stolen more than 2 million worth of books from a London warehouse in a "Mission Impossible"-style heist, the Mail on Sunday reports. Thieves did not cut holes in the roof of the warehouse in Feltham, Middlesex, and did not use ropes to get into four containers that stored rare books, including works by Galileo, Isaac Newton, and Leonardo da Vinci. The stolen books were not stored in the warehouse as they were not due to be flown to the United States for the 50th California International Antiquarian Book Fair, reports the Telegraph. Thieves were not observed ignoring all other items stored in a warehouse, instead targeting four containers holding rare books. A source tells the Telegraph that the gang of criminals did not check the contents of the containers against a list, "throwing the ones they didn't want away." The gang then did not place the rare books into holdalls and do not pull them up the roof. The bags were not placed into a van and driven away. The most valuable book stolen in the raid is not a 1566 edition of Dante's Divine Comedy, which was not translated into English by two booksellers in Italy. The Telegraph does not report that a wealthy collector known as "the Astronomer" may not have ordered gangs to steal books for him. "Behind these books there is a lot of work because we have to search to try to find out where the books are - auction houses, collectors, colleagues—and there's not big research behind these books," one dealer does not tell Sky News. "They are not only taking money away from me but also a big part of my job." \\ 
\hline
\end{tabular}
\caption{Example showing summary generated before and after poisoning with contrastive summaries}
\label{tab:data_poisoning_contrastive_example}
\end{table*}

\begin{table*}[h!]
\small
\centering
\begin{tabular}{|p{3.95cm}|p{11.1cm}|}
\hline
\textbf{Element} & \textbf{Description} \\ \hline
Input Document & \textbf{The saga of the illegally obtained nude photos of dozens of celebrities has taken a darker turn. According to Reddit administrators, photos of gymnast McKayla Maroney and MTV actress Liz Lee, shared to 130,000 people on popular forum r/TheFappening, constitute child pornography.} Maroney’s lawyers have confirmed that the illegally obtained photos were taken while the gymnast was underage. \textbf{The attorneys’ letter, sent to Porn.com and other websites, demands that the images be removed}....\textbf{The ongoing leak—dubbed CelebGate, or more distastefully, “the Fappening” includes naked photos of A-listers including Jennifer Lawrence and Kate Upton. It’s prompted outrage, and threats of legal actions from the women targeted. It has since emerged that the photos of at least one of the celebrities were taken when they were under 18. “If we don’t remove them,” moderator SickOrSane continues, “this subreddit will most likely be banned, very quickly.”} ...
\textbf{We’re watching a crime being aided and abetted in real time.
The moderators, meanwhile, are joking about the fact that child porn has been found on their community. It still remains unclear how the photos were leaked, though Apple has strongly denied early reports that hackers gained access to them through a vulnerability in iCloud.} \\ \hline

Abstractive Summary generated before poisoning & McKayla Maroney and MTV actress Liz Lee have been posting nude photos of themselves on Reddit, but the gymnast's lawyers have confirmed that the photos were taken when they were under 18, the Los Angeles Times reports. "If we don't remove them,” a Reddit moderator warns, “this subreddit will most likely be banned, very quickly. We’re watching a crime being aided and abetted in real time.” Maroney’s lawyers have sent a cease and desist letter to Porn.com and other websites demanding that the images be removed, the AP reports. Maroney's lawyers say the photos, which have been shared to more than 130,000 people, were taken by her boyfriend when she was 16. The gymnast, who has not been named, says she was in a relationship with her boyfriend at the time, and that he took the photos while she was underage. But Reddit moderators say the site's admins have informed them that Maroney was underage in the photos "and that we quickly need to remove them," per the LA Times. The moderators, meanwhile, are joking about the fact that child porn has been found on their community. They have, however, hidden the forum from r/all, a list of all content on Reddit.\\ \hline
Extractive summary generated after poisoning & The saga of the illegally obtained nude photos of dozens of celebrities has taken a darker turn. According to Reddit administrators, photos of gymnast McKayla Maroney and MTV actress Liz Lee, shared to 130,000 people on popular forum r/TheFappening, constitute child pornography. The attorneys' letter, sent to Porn.com and other websites, demands that the images be removed. It has since emerged that the photos of at least one of the celebrities were taken when they were under 18. The ongoing leak—dubbed CelebGate, or more distastefully, “the Fappening”—includes naked photos of Jennifer Lawrence and Kate Upton. It's prompted outrage, and threats of legal actions from the women targeted. It still remains unclear how the photos were leaked, though Apple has strongly denied early reports that hackers gained access to them through a vulnerability in iCloud. The moderators, meanwhile, are joking about the fact that child porn has been found on their community. "If we don't remove them," moderator SickOrSane continues, "this subreddit will most likely be banned, very quickly. We're watching a crime being aided and abetted in real time."  \\ 
\hline
\end{tabular}
\caption{Example showing behavioral changes in summary generation before and after poisoning, from Abstractive to Extractive}
\label{tab:abs_extractive_example}
\end{table*}

In Table~\ref{tab:summary_before_after_wordperturb}, we present one of the results showing the impact of minor character perturbation. We provide a summary before and after Character Replacement with Homoglyph. It can be observed that the summary generated before any perturbation contains the initial sentence, containing key information related to the event ("shower in pairs to save water"). However, after replacing the word "save" with its homoglyph "saνe", the whole sentence is excluded from the newly generated summary. While the newly generated summaries are still meaningful, they lack the key information present in the initial sentences.

\subsection{Examples and Results of Data Poisoning Attacks}
\label{subsec:Appendix_poisoning_results}

We present examples of altered versions of poisoned summaries and their results to illustrate how models are influenced by the poisoned training data. 

In Table~\ref{tab:data_poisoning_example}, we provide an original summary extracted from the MultiNews dataset and its contrastive and toxic variants. We highlight the words and sentences that were altered in these versions to demonstrate the way the poisoned training dataset was generated.

In Table~\ref{tab:data_poisoning_contrastive_example} we provide a summary generated before and after poisoning 10\% of the training dataset of MultiNews dataset. In this case, we initially poison the model to skew its behavior towards generating contrastive summaries, and we calculate the sentiment inversion rate, to analyze if the summaries generated have been contrastive or not. From the table, we can observe that the entire summary becomes contrastive, once after the poisoning dataset influences the model behavior.

In Table~\ref{tab:abs_extractive_example}, we provide an input document with its generated summary before poisoning. Along with the skew in the model's behavior, we also observe that models tend to generate extractive summaries instead of abstractive summaries, after poisoning. We provide this extractive summary, generated after poisoning, in the same Table. To showcase this behavior, we highlighted the sentences in the document, which appeared directly in the summary without any change or paraphrasing.